\documentclass{sig-alternate-10pt}
\usepackage{subfig}
\usepackage{multirow}
\usepackage{textcomp}

\newcommand{\textapprox}{\raisebox{0.5ex}{\texttildelow}}
\begin{document}

\title{Household Electricity Demand Forecasting \\- Benchmarking State-of-the-Art Methods}

\numberofauthors{1} 
\author{\alignauthor Andreas Veit, Christoph Goebel, Rohit Tidke, \\Christoph Doblander and Hans-Arno Jacobsen\\
\affaddr{Department of Computer Science, Technische Universit\"{a}t M\"{u}nchen} \\
\email{a.veit@in.tum.de, christoph.goebel@tum.de, rohittidke@gmail.com, \\doblande@in.tum.de, jacobsen@in.tum.de}
} 

\date{01 April 2014}

\maketitle
\begin{abstract}
The increasing use of renewable energy sources with variable output, such as solar photovoltaic and wind power generation, calls for Smart Grids that effectively manage flexible loads and energy storage. The ability to forecast consumption at different locations in distribution systems will be a key capability of Smart Grids. The goal of this paper is to benchmark state-of-the-art methods for forecasting electricity demand on the household level across different granularities and time scales in an explorative way, thereby revealing potential shortcomings and find promising directions for future research in this area. We apply a number of forecasting methods including ARIMA, neural networks, and exponential smoothening using several strategies for training data selection, in particular day type and sliding window based strategies. We consider forecasting horizons ranging between 15 minutes and 24 hours. Our evaluation is based on two data sets containing the power usage of individual appliances at second time granularity collected over the course of several months. The results indicate that forecasting accuracy varies significantly depending on the choice of forecasting methods/strategy and the parameter configuration. Measured by the Mean Absolute Percentage Error (MAPE), the considered state-of-the-art forecasting methods rarely beat corresponding persistence forecasts. Overall, we observed MAPEs in the range between 5 and >100\%. The average MAPE for the first data set was \textapprox 30\%, while it was \textapprox 85\% for the other data set. These results show big room for improvement. Based on the identified trends and experiences from our experiments, we contribute a detailed discussion of promising future research.  
\end{abstract}

\section{Introduction}
According to the US Department of Energy, the creation of a sustainable and energy-efficient society is one of the greatest challenges of this century, as traditional non-renewable sources of energy are depleting and adverse effects of carbon emissions are being felt~\cite{doe2003grid}. Two key issues in creating a sustainable and energy-efficient society are reducing peak energy demands and increasing the penetration of renewable energy sources. The authors of~\cite{goebel2014energy} outline a cs research agenda to help achieve this goal. To achieve a reliable operation of the electricity distribution system, supply and load have to be balanced within a tight tolerance in real time. Load forecasting has therefore been a major issue in power systems operations~\cite{141695}.Today, with increasing decentralized generation of electricity, there is a need for controlling of smaller zones of the electric grid. Smart Grids enable micromanagement of those zones. In~\cite{bunn1985review}, the authors describe how accurate load forecasts can greatly enhance the micro-balancing capabilities of smart grids, if they are utilized for control operations and decisions like dispatch, unit commitment, fuel allocation and off-line network analysis. Thus, the prediction of energy consumption is a vital factor towards successful energy management. 

Load forecasts can be performed on different voltage levels in the grid. Forecasts can be performed on the transmission level, the distribution level and even the individual household and device level, because with the introduction of smart meters, the load can now be measured on the household level. Even more granular forecasts can be performed on the appliance levels with installations of energy sensors or energy consumption disaggregation~\cite{ziekow2013potential}. Recently, there have been many studies on the disaggregation of electricity consumption of households into individual appliances~\cite{carrie2012disaggregation}. However, the short term forecasting of individual household consumption has not been evaluated to a satisfactory extent.

Considering the importance of short-term load forecasting in demand and supply balancing, we conduct experiments in order to compare state-of-the-art forecasting methods. The growing public availability of electricity consumption data gives the opportunity to analyze and benchmark possible forecasting methods and strategies. In our experiments we use Autoregressive Integrated Moving Average (ARIMA), exponential smoothing and neural networks for univariate time series. In addition, we apply three different forecasting strategies: a sliding window approach, a day type approach and a hierarchical day type approach. The analysis of these methods on multiple data sets can give an indication of the optimal parameterization and usage. Therefore, we used two electricity consumption data sets: one collected by researchers at the Technische Universit\"{a}t M\"{u}nchen and one from the Massachusetts Institute of Technology. For the comparison of the different methods and strategies we used different granularities of consumption data, i.e., sampling frequencies from 15 up to 60 minutes, and varied the time horizons for the forecasts from very short-term forecast of 15 minutes up to forecasts of 24 hours. In order to compare the results of the different methods and strategies and the influence of the granularity and forecast horizon, we tested the accuracy of the forecast with the Mean Absolute Percentage Error. Overall, we observed MAPEs in the range between 5 and >100\%, with the average MAPE for the first data set being \textapprox 30\% and \textapprox 85\% for the second data set, respectively. Looking at the performance of the algorithms and strategies, we see that most of the algorithms benefit from splitting the data into training sets of particular day types and that predictions based on disaggregated data from individual appliances leads to better results. Generally, we show that without further refinement of advanced methods such as ARIMA and neural networks, the persistence forecasts are hard to beat in short-term forecasts. Especially in households with demand profiles that remain constant for many hours during a typical day, advanced forecasting methods provide little value, if they are not embedded into a framework that adapts their use to individual household attributes. Therefore we also provide an exploration of promising directions for future research. These experimental results and the exploration of future research directions are the primary contributions of this paper.

This paper is organized as follows: In Section 2, we will review related literature and identify the research gap. In Section 3, we describe the electricity consumption data we use in our experiments and explain all performed transformations. Subsequently, in Section 4, we describe our experimental setup and the forecasting methods and strategies used in our experiments. Section 5 presents the results of our experiments and Section 6 discusses our findings and explores directions for future research.

\section{Related Work}
Demand side management and demand response receive increasing attention by research and industry. The research work published so far includes a variety of directions from direct load control or targeted customer interaction to indirect incentive-based control (see~\cite{medina2010demand} for an overview). In order to help balancing demand and supply, demand side management programs require accurate predictions of consumer demand.

The approaches for demand side management focus on different levels of the power system. On the grid operator level, studies focus for example on the minimization of power flow fluctuations~\cite{tanaka2011optimal} or the integration of renewable energy~\cite{wu2012wind}. The distribution grid operator uses consumption forecasts to balance grids with a high penetration of decentralized generation of renewable energy (e.g., \cite{kok2011dynamic}, \cite{goebel2013using}). Other studies look at the level of groups of consumers with a focus on game theoretic frameworks \cite{mohsenian2010autonomous} or virtual price signals~\cite{VeitX+13}. Most work on demand response, however, focuses on the level of end consumers. Recent research has studied the use of variable price signals for individual customers. These dynamic tariffs penalize consumption during certain periods of time with increased electricity prices, so that customers can respond by adjusting their consumption (e.g., \cite{alberini2011response}, \cite{herter2007exploratory}). However,~\cite{ramchurn2011agent} points out that demand side management with variable price signals can cause instabilities through load synchronization. To avoid uncontrolled behavior, accurate consumption forecasts can help utilities to select customers that are most suitable for a demand response program. 

First studies have analyzed the potential and first prototypes of consumption forecasts for individual households (e.g. \cite{ziekow2013potential}, \cite{ziekow2013forecasting} ). However, most work on household consumption focuses on disaggregation of electricity consumption. Examples include \cite{kolter2011redd}, \cite{kolter2011large}, \cite{humeau2013electricity}, \cite{akshay2013sustainable} and \cite{kleiminger2013occupancy}. The authors of~\cite{carrie2012disaggregation} give an overview of the state-of-the-art in this area. In this paper, we benchmark state-of-the-art forecasting models for household consumption and also evaluate how the disaggregation of consumption data influences the prediction of household consumption.

\section{Electricity consumption data}
The data collected by the smart meters or smart home infrastructures include differing sets of attributes. The most common metrics are wattage readings or accumulated energy at discrete time steps. While some consumption data sets are univariate time series only consisting of the overall electricity consumption reading from a household, other data sets consist of multivariate data including, for example, readings from a system of sensors distributed over a household. In our experiment we use data sets from the second category. 
\subsection{Data Sets}
We use two different data sets for our experiments. To perform the same experiments using both data sets, a transformation of the data was necessary. These transformations are explained in the following section. 
\begin{figure}
	\centering
	\includegraphics[width=\columnwidth]{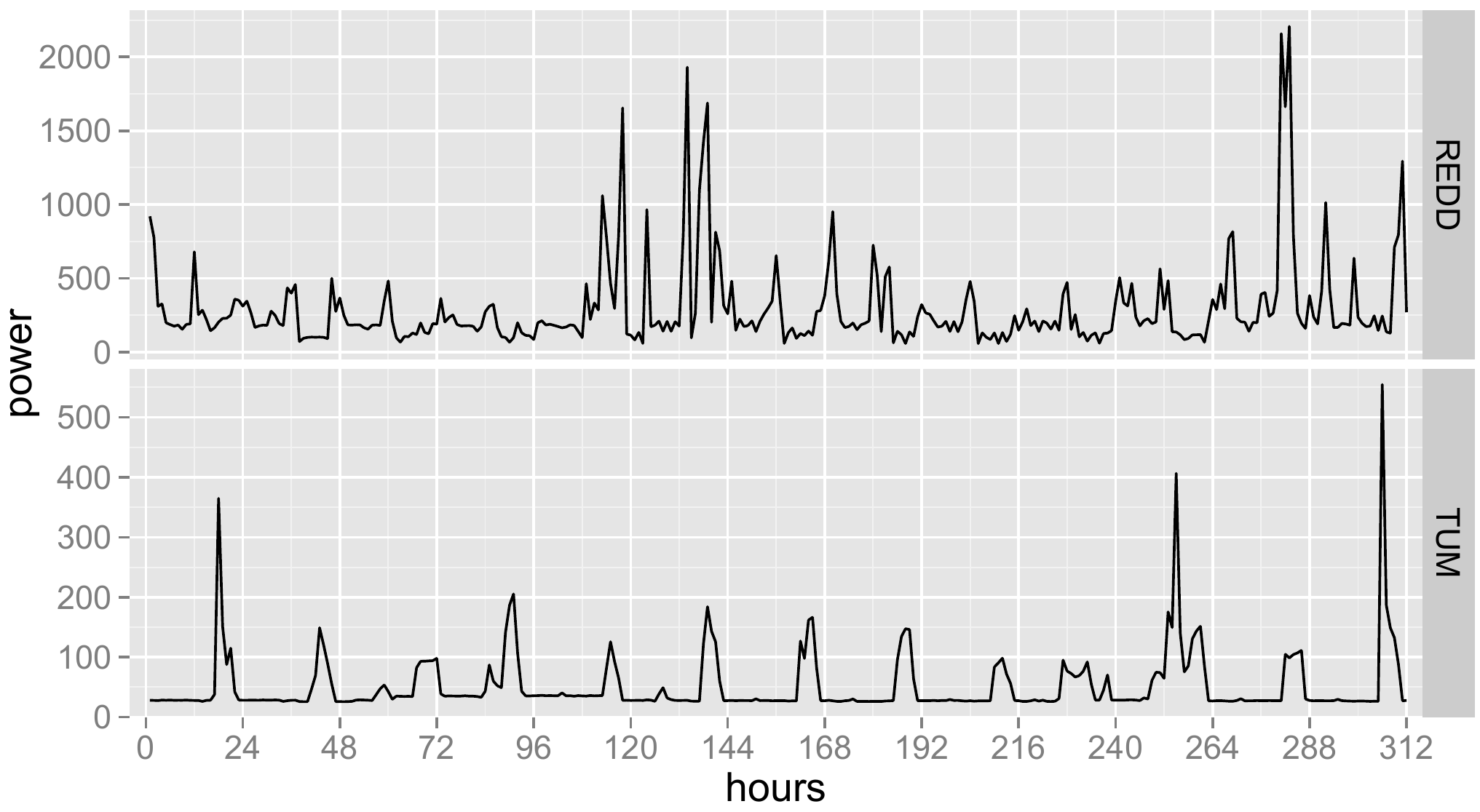}
    \caption{Demand profiles from the data sets.}
    \label{fig:profiles}
\end{figure}
\subsubsection{The TUM Home Experiment Data Set}
In the TUM Home Experiment, a single household in Germany, in the state of Bavaria, was equipped with a distributed network of Pikkerton sensors measuring power in Watt, on/off status and energy in kWh from several appliances. The measured appliances include lights in the kitchen, antechamber and living room, the fridge, washing machine, office and entertainment devices. The data used for this experiment was collected from February 4th 2013 to October 31st 2013. Figure~\ref{fig:profiles} shows in the lower graph the demand profile from February 21st to March 5th 2013. From the figure it can be seen that the demand is flat for long time intervals, with occasional peaks, especially in the evenings. In particular, 70\% of all power readings lie between 25 and 30W. Figure~\ref{fig:cumDist} shows the empirical cumulative distribution function of the power readings. The graph illustrates the very steep increase of power frequency at around 25 to 30 Watt. In the following, we will refer to this data set as the TUM data set.

\subsubsection{The Reference Energy Disaggregation Data Set}
\begin{figure}
	\centering
	\includegraphics[width=\columnwidth]{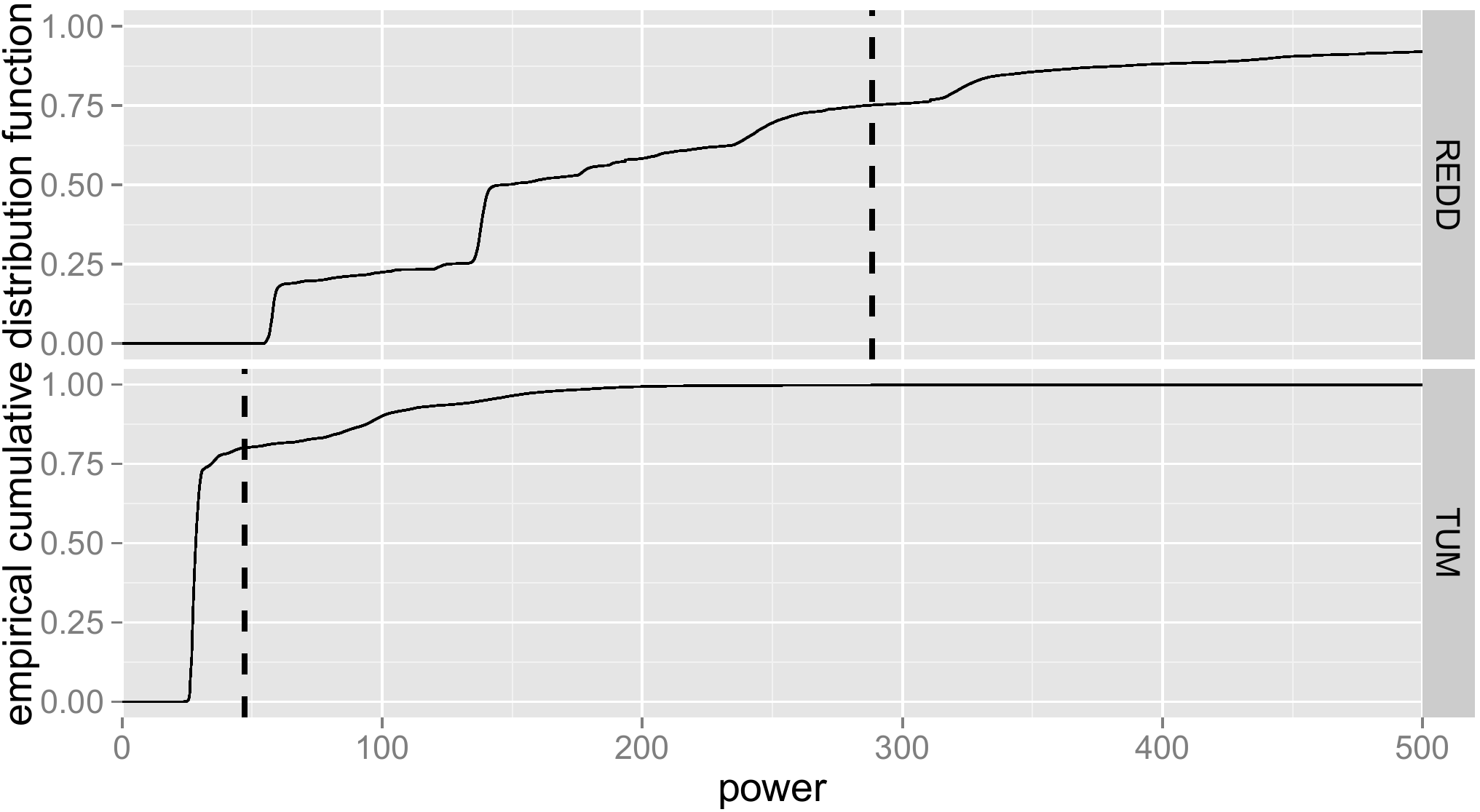}
    \caption{Cumulative distribution of power.}
    \label{fig:cumDist}
\end{figure}

The Reference Energy Disaggregation Data Set (REDD) is a public data set for energy disaggregation research~\cite{kolter2011redd}. The REDD data set is provided by the Massachusetts Institute of Technology and contains power consumption measurements of 6 US households recorded for 18 days between April 2011 and June 2011. The data set contains high frequency and low frequency readings and includes readings from the main electrical circuits as well as readings from individual appliances such as lights, microwave and refrigerator. For the experiments presented in this paper we use the low frequency readings of the individual appliances, which are sampled at intervals of 3 seconds. Figure~\ref{fig:profiles} shows in the upper graph the demand profile for house 1 from April 19th to May 1st 2011. From the figure it can be seen that, in contrast to the TUM data set, the REDD aggregate demand has more frequent and higher fluctuations. As a result, the cumulative distribution of the power readings shown in Figure~\ref{fig:cumDist} has a flatter slope. We will refer to this data set as REDD data set.
\newpage
\subsection{Data Transformation}
The data sets used in our experiment come in different formats. To achieve comparable results, they need to be transformed to obtain uniformity and to allow the generation of data sets at the required granularities for the experiment. The transformation can be divided into three steps:

Step 1: The data sets are transformed into a common format. Since the readings are at different frequencies, we convert the time indicators into UNIX timestamps and the granularity to one minute.

Step 2: Statistical time series forecasting relies on the assumption that time series are equally spaced. In~\cite{eckner2012algorithms}, the author explains that most research has been conducted on equally spaced time series. In~\cite{eckner2012framework}, he explains that in case of unequally spaced time series interpolation methods should be used to transform unequally spaced intervals into equally spaced intervals. Usually linear interpolations are performed for this transformation. After interpolating the gaps in the time series, standard models for equally spaced intervals can be used. In the data sets used for our experiments several breaks in the time series exist, due to meters or sensors not providing measurements. These breaks cannot all be interpolated, because the interpolation of long intervals can have a significant influence on the statistical forecasting model. As the main seasonality in electricity consumption data is one day, longer breaks of several hours can no longer be interpolated. Interpolation would otherwise distort the forecasting models. Determining the optimal length of interpolation intervals is itself an optimization problem. In this paper, we interpolate intervals up to a length of two hours using linear interpolation.

Step 3: The different strategies we use in our experiments require different formats for their data. First we use a sliding window strategy where we select training data windows of specific lengths to predict future load. For this strategy a continuous time series is necessary. Therefore we select the longest period without breaks longer than 2 hours. We also evaluate day type strategies, where the forecasting models are trained using data from similar days of the week. For these strategies we create a cross-sectional data set divided by the days of the week. We join each day of the week, e.g., Mondays of consecutive weeks, into one data set.

In the following, we explain the transformations performed on each data set.

\subsubsection{Transformation of TUM Data Set}
The TUM data set as introduced above is a multivariate data set containing measurements from several appliances in the experiment house. Table~\ref{tab:TUMdatasets} shows an extract from the raw data set. In order to be consistent with the REDD data set, the time stamps and the granularity are converted to the UNIX format and one minute intervals. 
Since the readings from all appliances have been stored in one big data set, we extract the power readings and split the data set into individual channels for the different appliances and subsequently interpolate gaps of up to two hours. Then, for the hierarchical strategy we created a separate data set for each appliance and for each day of the week. For several times, no data is available from some appliances, but data is available for other appliances. Such incomplete data could disrupt forecasts, because the forecasting model would assume the appliance to be switched off, although it is running. To obtain a consistent data set, we only consider durations where data is available from all appliances. Afterwards, we aggregate the different appliance channels for the day type and the sliding window strategies. To get a continuous time series for the sliding window strategy, we choose to only use the data from the longest consistent distinct data set, which is the data of the period from Feb 20th 2013 09:13:00 GMT to Apr 5th 2013 05:44:00 GMT.

\begin{table*}
\centering
\caption{An extract from the TUM data set.}
\label{tab:TUMdatasets}
\begin{tabular}{|c|c|c|c|c|}
	\hline
	timestamp & value & property & unit & appliance\tabularnewline
\hline 
\hline 
2013-10-01 20:21:33 & 1.538 & WORK & kWh & light-livingroom\tabularnewline
\hline 
2013-10-01 20:21:33 & ON & POW & boolean & light-livingroom\tabularnewline
\hline 
2013-10-01 20:21:33 & 501875 & FREQ & Hz & light-livingroom\tabularnewline
\hline 
2013-10-01 20:21:33 & 231 & VRMS & RMS & plug-office\tabularnewline
\hline 
2013-10-01 20:21:33 & 30 & LOAD & Watt & washingmachine\tabularnewline
\hline 
2013-10-01 20:21:34 & 0 & LOAD & Watt & washingmachine\tabularnewline
\hline 
2013-10-01 20:21:34 & 55 & IRMS & RMS & washingmachine\tabularnewline
\hline 
2013-10-01 20:21:34 & 0.636 & WORK & kWh & washingmachine\tabularnewline
\hline 
2013-10-01 20:21:34 & ON & POW & boolean & washingmachine\tabularnewline
\hline 
2013-10-01 20:21:34 & 49.7500 & FREQ & Hz & washingmachine\tabularnewline
\hline 
2013-10-01 20:21:34 & 231 & VRMS & RMS & washingmachine\tabularnewline
\hline
\end{tabular}
\end{table*}

\subsubsection{Transformation of REDD Data Set}
The REDD data set also contains measurements from several appliances. They are already divided into separate channels for the individual appliances. To be consistent with the TUM data set the time stamps and the granularity have been converted to the common UNIX format. Although the data set contains readings from six different houses, we only used the data from house no. 1, as it contains a long enough period of measurements. For the other houses we can neither perform the day type nor the hierarchical forecasting strategies, as they only contain data from 2 up to 3 days for each day of the week. For the data from house 1 we then interpolate gaps of up to two hours and aggregate the different appliance channels for the day type and the sliding window strategies. To get a continuous time series for the sliding window strategy, we choose the longest consistent distinct data set, i.e., the data from Apr 18th 2011 22:00:00 GMT to May 2nd 2011 21:59:00 GMT.

\section{Experiments}
In this section we will introduce the different forecasting methods and strategies we use in our experiment as well as their specific parameterizations. 
\subsection{Forecasting Methods}
First, as a benchmark for the other forecasting methods we include the persistence method, where all forecasts are equal to the last observation. We will refer to this method as PERSIST. For short forecasting horizons and high granularities of consumption data, persistence forecasts are known as hard to beat by the other methods. 

Furthermore, we use the Autoregressive Integrated Moving Average (ARIMA) model. The model is denoted as ARIMA$(p,d,q)(P,D,Q)$, where the non-seasonal components are defined in the first parentheses and the seasonal components of the model are defined in the second parentheses. The parameters $\left(p,P\right)$ denote the number of lagged variables, i.e., the number of last observations for autoregression in the seasonal and non-seasonal components. The parameters $\left(d,D\right)$ denote the difference that is necessary to make the time series stationary. Lastly the parameters $\left(q,Q\right)$ denote the moving average over the number of last observations. To find the optimal parameters, we use the \texttt{auto.arima()} method provided by the \texttt{R forecast} package. This function provides the best ARIMA model according to the minimization of Akaike information criterion with a correction for finite sample sizes(AICc). The algorithm to determine the model parameters is described in~\cite{hyndman2007automatic}.

Third, we use an Exponential smoothing state space model (BATS) with Box-Cox transformation, ARMA errors as well as trend and seasonal components. The model is denoted as BATS$\left(\omega, \phi, p, q, m_{1}, m_{2}...m_{t}\right)$, where $\omega$ is the Box-Cox, $\phi$ the damping, $\left(p, q\right)$ the ARMA parameters, and $\left(m_{1}, m_{2}...m_{t}\right)$ are the seasonal periods. We also apply the TBATS model, which uses trigonometric functions for the seasonal decomposition. It is denoted as TBATS$\left(\omega, \phi, p, q, \{m_{1}k_{1}\}, \{m_{2}k_{2}\}...\{m_{t}k_{t}\}\right)$, where the parameter $k_{i}$ represents the number of harmonics required by the $i$th seasonal component. The BATS and TBATS approach is explained in detail in~\cite{de2011forecasting}. We used the implementations of the \texttt{bats()} and \texttt{tbats()} functions provided by the \texttt{R forecast} package. 

Lastly, we also use feed-forward neural networks with a single hidden layer and lagged inputs for forecasting univariate time series. The model is denoted as NNAR$\left(p,P,k\right)_{m}$, where $p$ is the number of non-seasonal lags, $P$ is the number of seasonal lags, $k$ is the number of nodes in the hidden layer and $m$ is the seasonal period. The model is analogous to an ARIMA$\left(p,0,0\right)\left(P,0,0\right)$ model, but with nonlinear functions. We used the implementations of the \texttt{nnetar()} function provided by the \texttt{R forecast} package. In this function the network is trained for one-step forecasts. For forecasts of longer horizons, forecasts are computed recursively. 
\subsection{Forecasting Strategies}
In our experiment we use three different strategies to sample the training and test data. In the following we introduce the individual strategies:
\begin{figure}
\centering
\subfloat[Sliding Window Strategy]
{
\label{fig:windowing}
\includegraphics[width=\columnwidth]{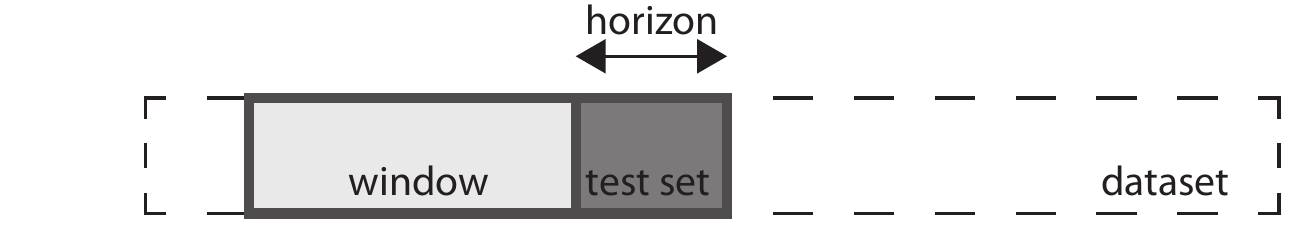}
}\\
\subfloat[Day Type Strategy]
{
\label{fig:daywise}
\includegraphics[width=\columnwidth]{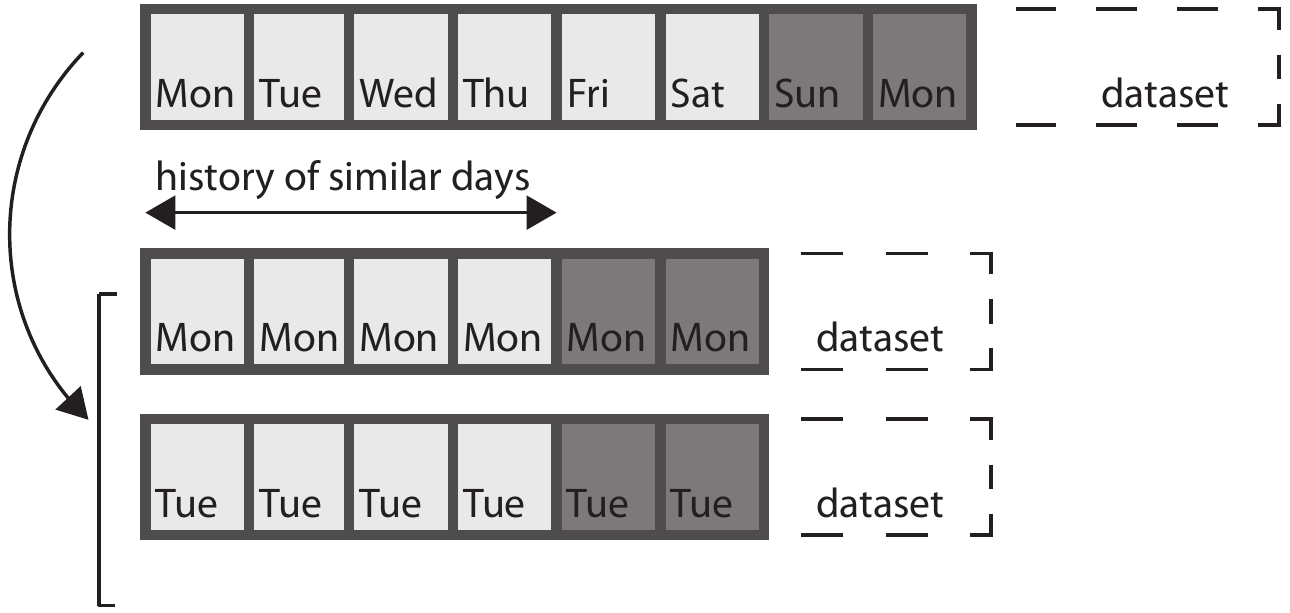}
}\\
\subfloat[Hierarchical Day Type Strategy]
{
\label{fig:hierarchical}
\includegraphics[width=\columnwidth]{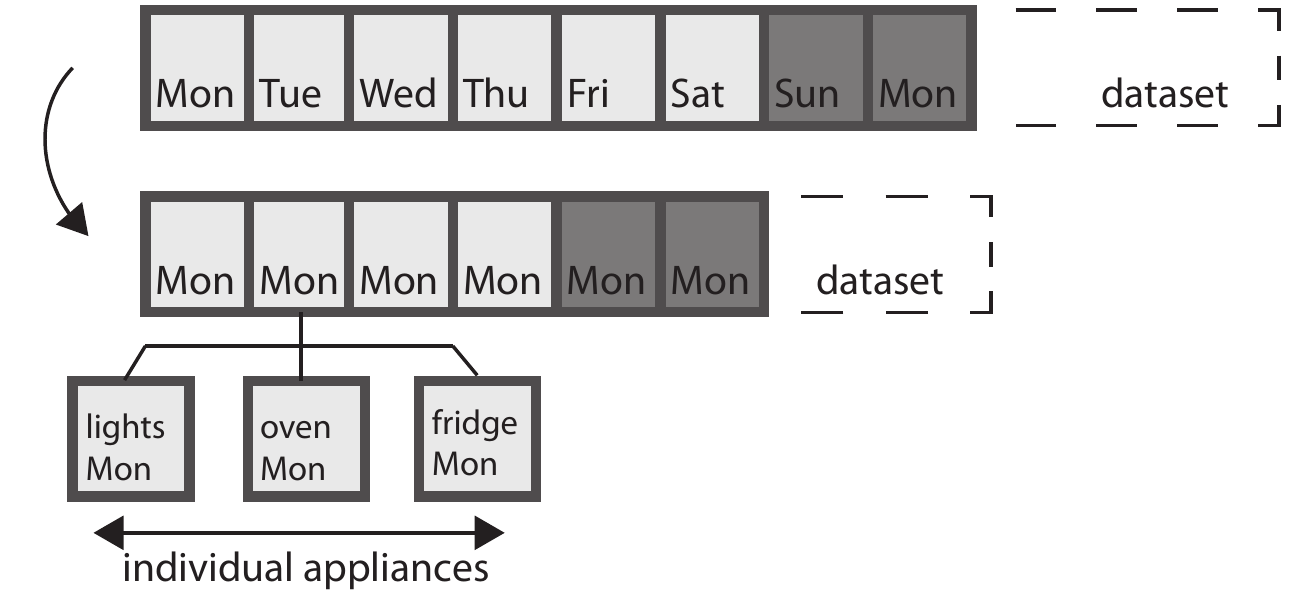}
}
\caption{Different forecasting strategies.}
\label{fig:strategies}
\end{figure}
\subsubsection{Sliding Window Strategy}
First, we used the \emph{sliding window strategy}, where the data set is divided into windows of smaller parts with a defined length. Each created window of training data has a corresponding test windows for cross validation to measure the accuracy of the prediction. The approach is illustrated in Figure~\ref{fig:windowing}. The forecasting model is then fitted to the validated window and tested against the test window. The main reason for using the sliding window approach is that the available data sets are of different length. Using standardized window lengths allows comparing the results from different data sets. After a prediction model has been trained and tested, the window moves forward on the data set. The distance the window is moved is called sliding length. In our experiment we use sliding windows with a sliding length of 24 hours.  
\subsubsection{Day Type Strategy}
Second, we use a \emph{day type strategy}. While the sliding window approach considers the data to be a continuous time series, the day type approach uses cross-sectional data. The strategy is to join each day of the week of consecutive weeks into separate data sets. The approach is illustrated in Figure~\ref{fig:daywise}. The training data set and the test data set are then sampled from the individual data sets. An example of such an approach is to join the Mondays of consecutive weeks. 
\subsubsection{Hierarchical Day Type Strategy}
Third, we use a \emph{hierarchical day type strategy}. A hierarchical time series is a collection of several time series that are linked together in a hierarchical structure. Hierarchical forecasting methods allow the forecasts at each level to be summed up in order to provide a forecast for the level above. Existing approaches to hierarchical time series include a top-down, bottom-up and middle-out approach. In the top-down approach the aggregated series is forecasted and then disaggregated based on historical proportions. The possible ways to compute these proportions are explained in [27]. The bottom-up approach, first forecasts all the individual channels on the bottom level and then aggregates the forecasts to create the aggregated forecast. The middle-out approach combines both approaches. It starts at a middle layer or an intermediate level and uses aggregation for the higher layers and disaggregation for the lower layers. 
We apply a bottom-up approach, i.e., we use the individual appliance channels to create forecasts for the individual appliances. Similar to the day type approach, we join each day of the week of consecutive weeks into separate data sets for each appliance. The approach is illustrated in Figure~\ref{fig:hierarchical}. Finally, we aggregate the individual forecast to a forecast of the entire household and test it against the test window. 

\subsection{Granularities}
While both data sets used in this experiment contain data at 1-3 seconds granularity, other data sets and meters offer measurements of different granularities. Therefore, we want to understand the effect of different measurement granularities on the performance of the different forecasting methods. In particular, we transformed the available data into granularities of 15, 30 and 60 minute intervals. The power reading of those intervals is defined as the mean power of the readings in the respective interval.

\subsection{Training Window Sizes}
Another parameter for demand forecasting is the window length used for training the models. The different data sets and forecasting methods limit the length of training sets. For example, for the day type strategy only training sets of 3 days can be used, since the REDD data set only contains only four days for of each day of the week. The \texttt{ARIMA} method of the \texttt{R forecast} package cannot handle models with seasonal periods with more than 350 data points and the \texttt{NNET} method requires at least two seasonal period cycles to train the neural network. Considering these restrictions, we use a training window size of 3 days for the day type and the hierarchical day type approach and varied the training window length for the sliding window approach between 3, 5 and 7 days. 

\newpage
\subsection{Forecasting Horizons}
The forecasting horizon is the number of point forecasts the particular algorithm predicts into the future. In the context of this experiment the horizon is given by the minutes the load is predicted into the future. The focus of this work lies on short-term forecasts. Hence, the range of the prediction lies between 15 minutes and 24 hours. Note that the granularity of the forecast cannot be higher than the granularity of the training data. For instance, with a training data set of 15 minutes intervals the earliest prediction will be 15 minutes into the future and all further predictions will be in intervals of 15 minutes. 

\subsection{Model Quality Measure}
We require a statistical quality measure that can compare the different forecasting methods and strategies. The present experiment uses the Mean Absolute Percentage Error (MAPE) as the standard accuracy error measure. The reason for this choice is that MAPE can be used to compare the performance on different data sets, because it is a relative measure. MAPE is defined as the mean over the ratio of the absolute difference between the residual and the actual value in percent: 
\begin{equation*}
\text{MAPE }=\frac{1}{n}\sum_{t\text{=1}}^{n}\lvert\frac{x_{t}-\hat{x}_{t}}{x_{t}}\rvert
\end{equation*}
where $x_{t}$ is the actual value and $\hat{x}_{t}$ is the forecast value. For example, with an actual load of 100 Watt and a corresponding forecasted load of 150 Watt, the MAPE would be 50\%, because the difference between actual and predicted load is 50\% of the actual load.

\subsection{Experimental Setup}
The purpose of our experiment is to gain insights into how the different parameters influence the different forecasting methods and strategies. This information helps to choose the most appropriate method. The following gives a summary of the different parameters and their values, as we used them in our experiment.
\begin{equation*}
\begin{split}
&granularity\in\left\{ 15, 30, 60\right\}\text{minutes}, \\
&method \in\left\{ARIMA, BATS, NNET, PERSIST, TBATS\right\}\\
&strategy \in\left\{ day type, hierarchical, sliding window\right\}\\
&horizon \in\left\{ 15, 30, 60, 180, 360, 720, 1440\right\} \text{minutes} \\
&window size \in\left\{ 3, 5, 7\right\} \text{days}
\end{split}
\end{equation*}

\section{Experimental Results}
In this section we present the influence of the defined parameters on the accuracy of the forecasting methods and strategies. For the evaluation we performed a total of 16038 different forecasts.

\emph{Result 1: For certain households increasing training window sizes significantly improve forecast accuracy.}

\begin{figure}
	\centering
	\includegraphics[width=\columnwidth]{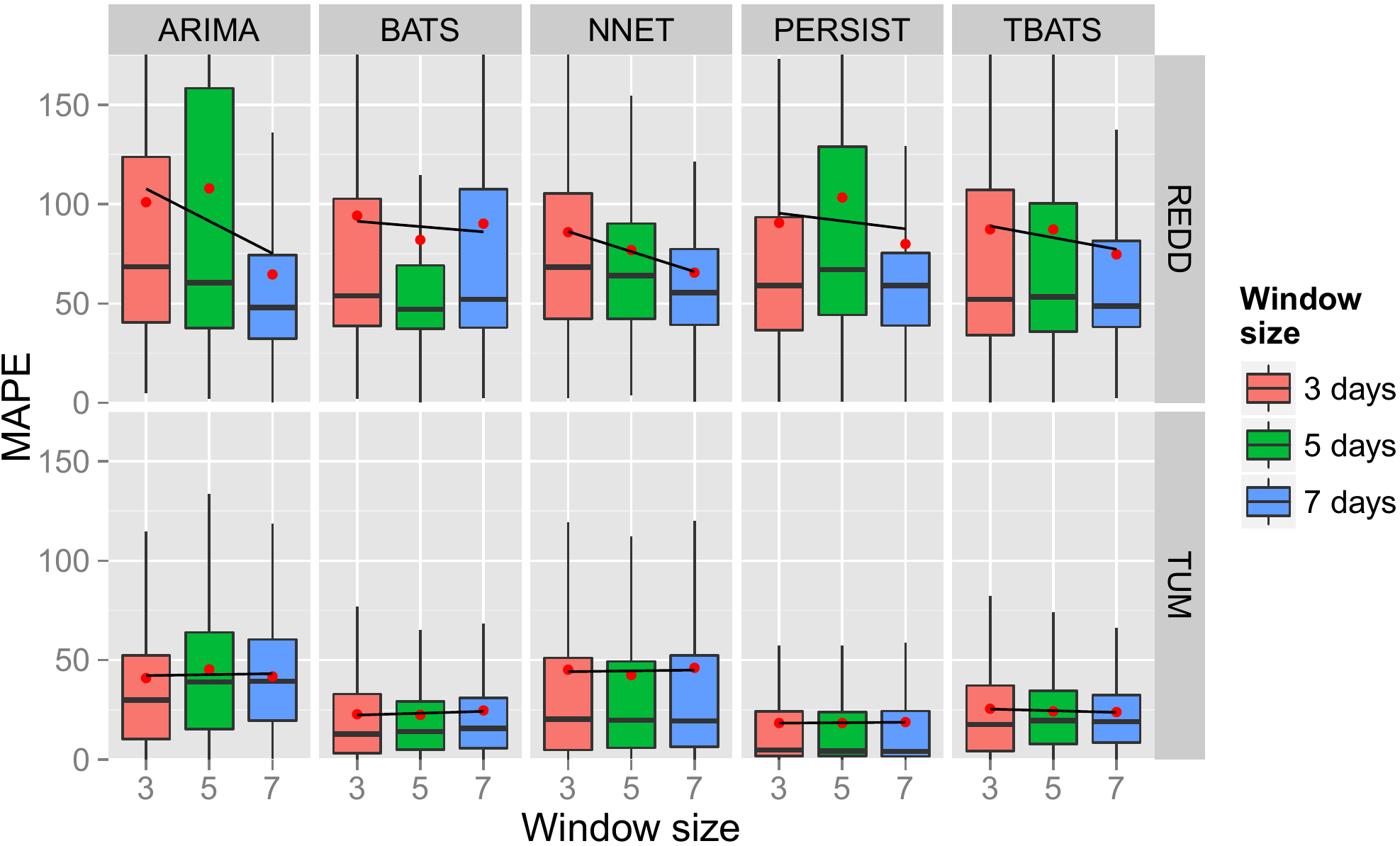}
    \caption{MAPE for varying window sizes.}
    \label{fig:window}
\end{figure}
\noindent
Figure~\ref{fig:window} shows boxplots of the distribution of the MAPE for the sliding window strategy and the different window lengths. The results are split by data set as well as by applied forecasting method. For each window length one boxplot shows the median as well as 25 and 75\% quantiles of the MAPE. In addition, the graph shows the mean as dots and a linear trend line over the increasing window sizes. From these results we have three key insights: (1) Increasing window sizes reduce the forecasting error on the REDD data set significantly, F(2, 2399)=10.209, p<0.0001. However, the forecasting error on the TUM data set does not change with increased window sizes F(2,11380)=0.1563, p>0.05. A possible explanation is that in the consumption profile of the TUM data set the consumption of every day is very similar and shows a constant pattern. Thus, an additional day of training data does not provide the models with new important information. However, on the REDD data set with its fluctuations in the demand profile the results of the ARIMA, NNET and TBATS methods can improve with the additional information. (2) The forecasting error on the TUM data set is almost constantly lower than on the REDD data set. This could be due to the fact that the demand profile of the TUM data set has long and frequent periods of constant consumption, which are easier to predict. The REDD data set on the other hand contains more fluctuations. (3) On the TUM data set, the persistence forecast has a better precision than all other forecast methods. This is also due to the long periods with constant consumption.

\emph{Result 2: Longer forecasting horizons lead to increasing errors. Lower granularities reduce the error.}

\begin{figure}
\centering
\subfloat[Sliding Window Strategy]
{
\label{fig:win}
\includegraphics[width=\columnwidth]{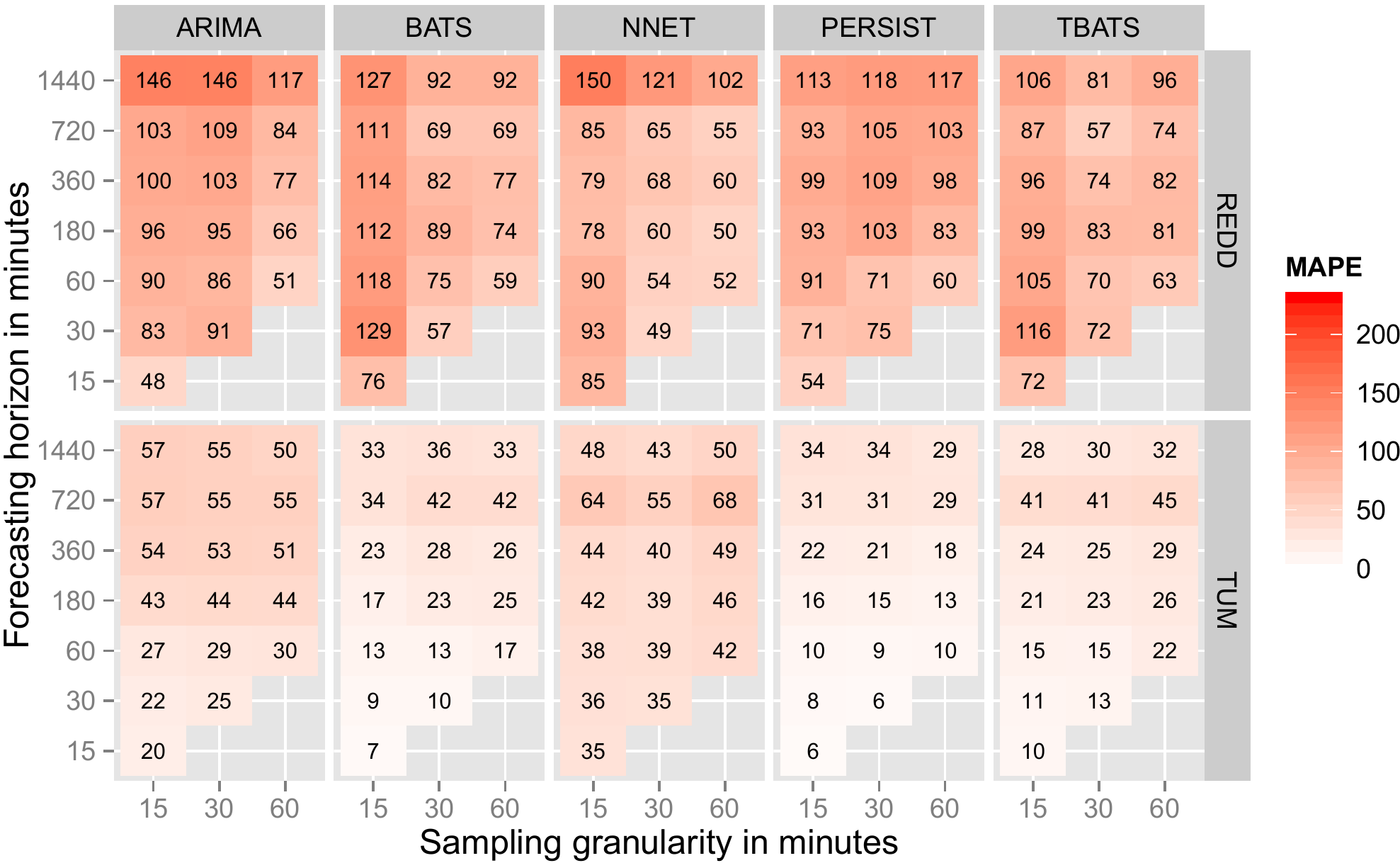}
}\\
\subfloat[Day Type Strategy]
{
\label{fig:day}
\includegraphics[width=\columnwidth]{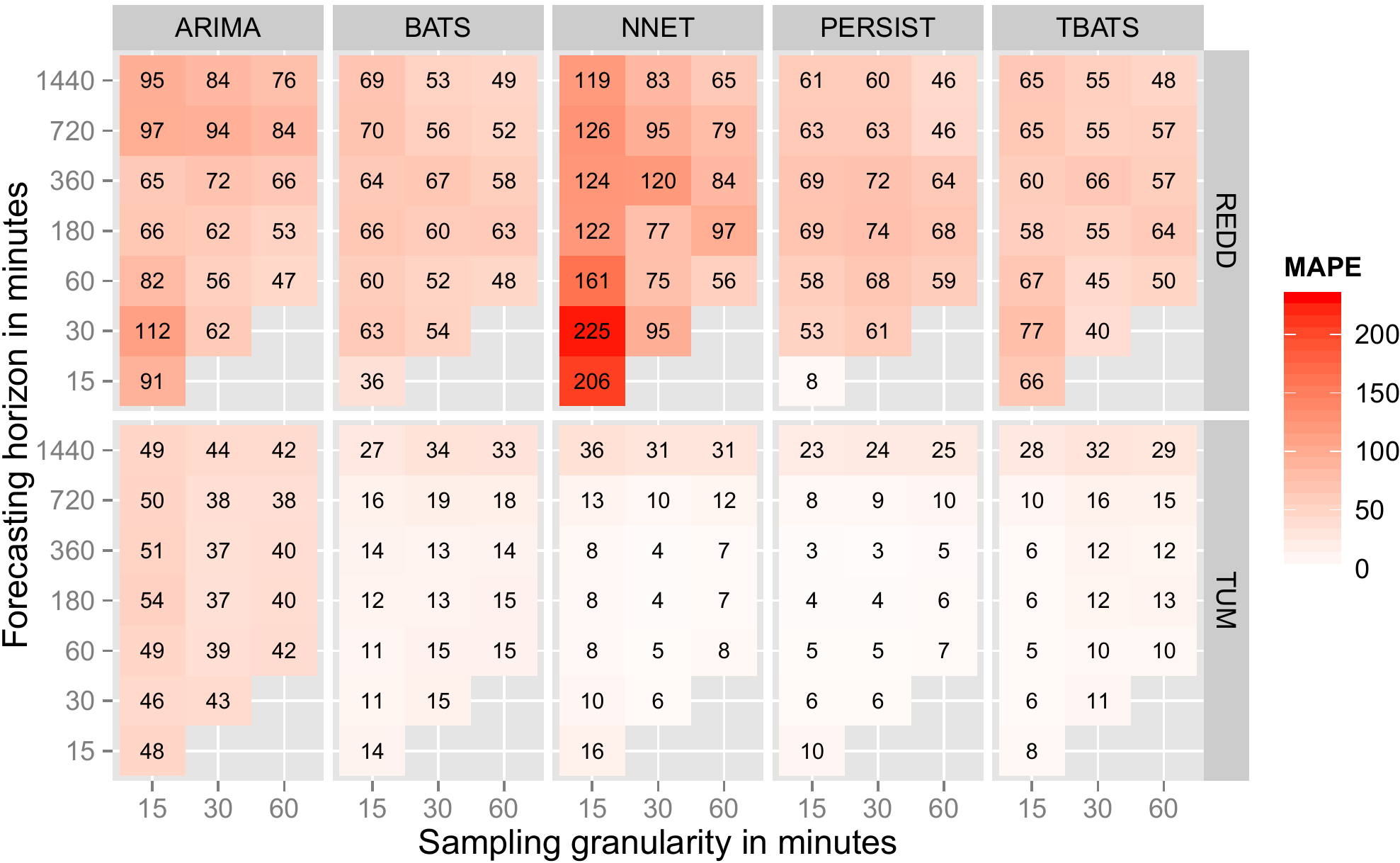}
}
\caption{MAPE for varying horizons and granularities.}
\label{fig:granHor}
\end{figure}
\noindent
Figure~\ref{fig:granHor} shows heatmaps of the MAPE from the sliding window and day type strategies for different granularities and forecasting horizons. The results from the hierarchical strategy are not included, because only the ARIMA method was performed for the hierarchical strategy. The results are split by data set as well as by the performed forecasting method. The values in the lower right corner of the respective tables are missing, because in these cases no forecast is possible, as the forecasting horizon is shorter than the data set granularity. From these results we gain four key insights: (1) With the exception of neural networks on the REDD data set, all forecasting methods can achieve better results on both data sets with the day type strategy. (2) Again, except for the neural network with the day type strategy on the REDD data set, longer forecasting horizons lead to larger errors. However, it can be observed that the exponential smoothing methods BATS and TBATS are more robust against increasing horizons than the other methods. (3) Especially on the REDD data set it can be observed that higher granularities lead to smaller errors. This can be explained by the reduction of the variance due to averaging the power readings over longer time intervals. With less variation in the demand profile the forecasting methods can make more precise predictions. As the demand profile of the TUM data set is constant over long periods, the increasing granularity does not decrease the error. (4) While in the REDD data set the persistence method has a high precision for short horizons and granularities, the exponential smoothing strategies BATS and TBATS and the neural network outperform the persistence method for granularities of 30 and 60 minutes. As mentioned above, the persistence forecast is difficult to beat on the TUM data set, but the exponential smoothing strategies BATS and TBATS get close, especially for longer forecasting horizons.

\emph{Result 3: For some households the prediction on weekdays can reach a higher precision than for weekends.}

\begin{table}
\centering
\caption{Distribution of power on weekdays and -ends.}
\label{tab:wdwe}
\begin{tabular}{|c|c|c|c|c|}
\hline 
 & \multicolumn{2}{c|}{TUM data set} & \multicolumn{2}{c|}{REDD data set}\tabularnewline
\hline 
\hline 
 & weekday & weekend & weekday & weekend\tabularnewline
\hline 
mean & 44.0 & 55.3 & 259.5 & 389.1\tabularnewline
\hline 
sd & 61.1 & 86.6 & 375.7 & 661.0\tabularnewline
\hline 
\end{tabular}
\end{table}
\begin{figure}
	\centering
	\includegraphics[width=\columnwidth]{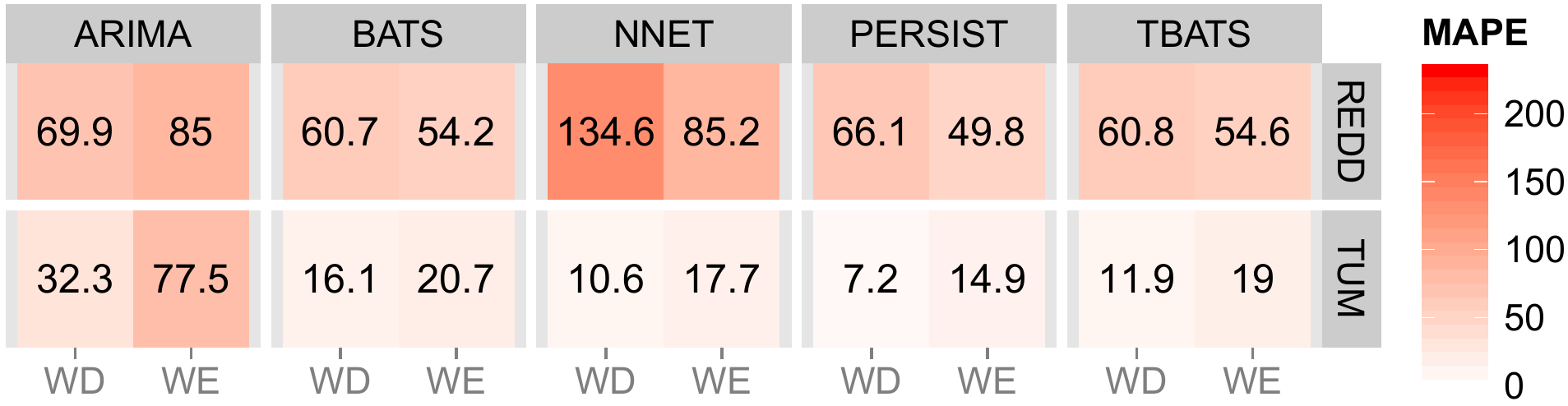}
    \caption{Mean MAPE for weekdays and -ends.}
    \label{fig:WDWE}
\end{figure}
\noindent
Figure~\ref{fig:WDWE} shows the mean MAPE from the day type strategy for weekdays and weekends. The results are split by data set as well as by the performed forecasting method. From these results can see that on the TUM data set all methods perform better on weekdays. However, on the REDD data set only the ARIMA model performs better on weekdays than on weekends, while all other methods perform better on the weekend. Table~\ref{tab:wdwe} shows the mean power consumption and the standard deviation of the power on weekdays and weekends for both data sets. For both data sets the mean consumption as well as the standard deviation are lower on weekdays than on weekends. On the REDD data set all the forecasting methods except ARIMA do not seem to be able to improve their precision with the reduced deviation. 

\emph{Result 4: Splitting the data set into day type windows can improve the forecast precision. In addition splitting the data set into distinct channels for individual appliances can also improve forecast precision.}

\begin{figure}
	\centering
	\includegraphics[width=\columnwidth]{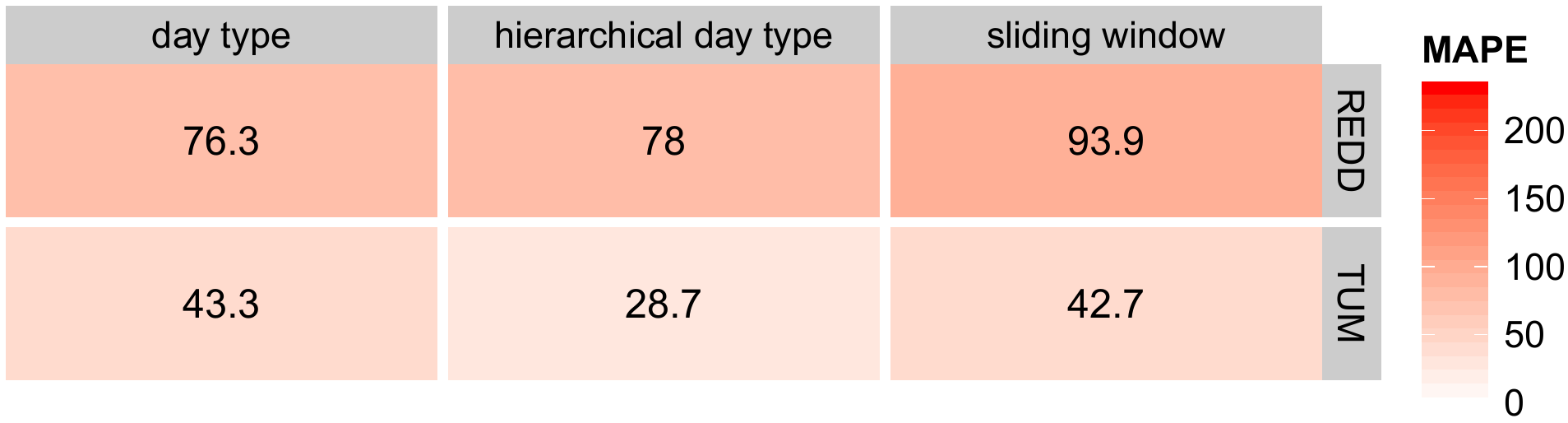}
    \caption{Mean MAPE for different strategies.}
    \label{fig:strat}
\end{figure}
\noindent
Figure~\ref{fig:granHor} shows that for almost every method a division of the data into day type windows improves the forecast precision against using simple sliding windows. In addition, Figure~\ref{fig:strat} shows a comparison of the mean MAPE of all three strategies for the ARIMA method. While the ARIMA method does not provide the best results in general, the figure shows that using a hierarchical strategy can greatly improve the performance on the TUM data set. This is a surprising result as generally the prediction of aggregated loads tend to result in a higher precision.

\section{Discussion}
Forecasting electricity consumption at different locations in electric distribution grids on short time scales is a crucial ingredient of systems that will enable higher renewable penetration without sacrificing the security of electricity supply. The goal of this paper is to evaluate the performance of state-of-the-art forecasting methods based on actual data. Overall, we observed that most of the algorithms benefit from larger training sets and splitting the data into training sets of particular day types. In addition we observed that predictions based on disaggregated data from individual appliances lead to better results. Generally, our analysis has revealed that if the forecasting methods are applied without individual tuning, they are able to beat the accuracy of persistence forecasting only in rare cases. Furthermore, the achievable accuracy in terms of average MAPE is surprisingly low, ranging between 5 and 50\% for one of the considered data sets, and between 30 and 150\% for the other, more variable, demand profile. Our work thus motivates more research investigating how accuracy can be increased.

R is only one out of many statistical packages offering state-of-the-art forecasting methods. As mentioned above, its data processing capabilities are limited. Other well-known packages that could be used include WEKA time series forecasting~\cite{Weka.2009} as well as several Python modules, e.g., statsmodel~\cite{statsmodels.2010} and Scikit-learn~\cite{scikit-learn}. 
Using Python modules, allows for fitting models to more data points compared to R and could therefore yield better results.

Furthermore, the introduction of further features could provide additional information for prediction algorithms to react faster in case a change in consumption occurs.
For example, when a device is switched on or off it takes some time until the average wattage of the time interval accounts for the change. The TUM data set contains additional sensors which are not yet considered in our experiments. We expect an increased forecast precision when information from occupancy, temperature and brightness sensors are included. 

We also expect error reduction when the consumption patterns of the appliances itself are considered. This is supported by the results for the hierarchical strategy (cf, Figure~\ref{fig:strat}). Thermal devices like fridges, freezers, boilers and heat pumps have a very predictable consumption pattern. Other devices like washing machines, dishwashers and laundry dryers have a known consumption pattern once switched on.
When looking at individual appliances, another direction worthwhile investigating would be event detection. Instead of prediction solely based on continuous wattage readings it could be beneficial to detect concrete events (e.g., on/off) and based on that derive a future consumption pattern. A sequence of events could train a markov model~\cite{Mathusamy.2010} and predict future events which could be used for the consumption forecast. We think that this could reduce the prediction error especially for short time forecasts.

In addition, it is important to investigate strategies for handling missing sensor data. In our experiments we only considered consistent data sets. However, in a real world setting load forecasts need to be performed even in situations with missing data. Future work should investigate how to handle temporary sensor outages, which could distract the prediction algorithms. 
 
Our results show a large difference between the forecasting accuracy of the same methods applied to two different data sets. It is unclear how common the characteristics of these data sets are. The necessary data for carrying out more representative studies is currently missing, although more data sets are currently being published~\cite{barker.2012}.
Since we tested a wide range of combinations of methods, strategies, sampling granularities and forecasting horizons, our experimental results give a good insight into how the different methods and strategies perform in various settings, despite the large difference in forecast accuracy.

In summary, this study should be considered as an exploration of promising directions for future research rather than yielding final results on the viability of local electricity demand forecasting.

\section{Conclusions}
We have evaluated a wide range of state-of-the-art methods and strategies for short-term forecasting of household electricity consumption, which is a key capability in many smart grid applications. Although our current data base is limited, we were able to gain useful insights into their performance at different levels of granularity and forecasting horizon length. We showed that without further refinement of advanced methods such as ARIMA and neural networks, the persistence forecasts are hard to beat in most situations. Especially in households with demand profiles that remain constant for many hours during a typical day, advanced forecasting methods provide little value, if they are not embedded into a framework that adapts their use to individual household attributes. Future work will focus on the design of such frameworks and evaluate them based on representative data. 

\scriptsize
\bibliographystyle{abbrv}
\bibliography{refs_energy}

\end{document}